%% file: root.tex
\documentclass[letterpaper, 10 pt, conference]{ieeeconf}  

\IEEEoverridecommandlockouts                              

\usepackage{graphics} 
\usepackage{mathptmx} 
\DeclareMathAlphabet{\mathcal}{OMS}{cmsy}{m}{n}
\usepackage{cite}

\usepackage{times}
\usepackage{epsfig}
\usepackage{graphicx}
\usepackage{amsmath}
\usepackage{amssymb}
\usepackage{color, colortbl}

\usepackage{helvet}
\usepackage{courier}

\usepackage{url}
\usepackage{rotating}
\usepackage{xspace}
\usepackage{booktabs}       
\usepackage{nccmath}
\usepackage{amsfonts}       
\usepackage{nicefrac}       
\usepackage{microtype}      
\usepackage{comment}		
\usepackage{adjustbox}		
\usepackage{tabu}
\usepackage{calc}			
\usepackage{wrapfig}

\usepackage{multirow}
\usepackage{float}
\usepackage[hang,flushmargin]{footmisc}
\usepackage[neverdecrease]{paralist}
\usepackage{caption}
\usepackage{subcaption}
\usepackage{algorithm}
\usepackage{algorithmic}
\usepackage{calc}
\usepackage{color}
\usepackage[font=footnotesize]{caption}
\usepackage{pifont}
\usepackage[dvipsnames]{xcolor}
\usepackage{makecell}
\usepackage{array}
\usepackage{hyperref}

\include{def2}

\definecolor{LightCyan}{rgb}{0.88,1,1}
\definecolor{White}{rgb}{1,1,1}
\definecolor{CuGray}{gray}{0.95}

\newcommand{\figref}[1]{Fig.~\ref{#1}}
\newcommand{\tabref}[1]{Table~\ref{#1}}

\newcommand{\secref}[1]{Sec.~\ref{#1}}

\newcommand{\splitcell}[1]{\begin{tabular}{@{}l@{}}#1\end{tabular}}
\newcommand{\bsplitcell}[1]{$\left[\splitcell{#1}\right]$}

\makeatletter
\DeclareRobustCommand\onedot{\futurelet\@let@token\@onedot}
\def\@onedot{\ifx\@let@token.\else.\null\fi\xspace}

\def\eg{\emph{e.g}\onedot}

\title{\LARGE \bf
Correlate-and-Excite: Real-Time Stereo Matching via\linebreak 
Guided Cost Volume Excitation
}

\author{Antyanta Bangunharcana$^{1}$, Jae Won Cho$^{2}$, Seokju Lee$^{2}$, In So Kweon$^{2}$, Kyung-Soo Kim$^{1}$, Soohyun Kim$^{1}$ 
\thanks{$^{1}$A. Bangunharcana, K-S. Kim, and S. Kim are with Mechatronics, Systems and Control Laboratory, KAIST, Daejeon, 34141, Republic of Korea {\tt\small \big\{antabangun, kyungsoo, soohyun\}@kaist.ac.kr}}%
\thanks{$^{2}$J. W. Cho, S. Lee, I. S. Kweon are with the Robotics and Computer Vision Laboratory, KAIST, Daejeon, 34141, Republic of Korea {\tt\small \big\{chojw, seokju91, iskweon77\}@kaist.ac.kr}}%
}

\begin{document}

\maketitle
\thispagestyle{empty}
\pagestyle{empty}

\begin{abstract}
    Volumetric deep learning approach towards stereo matching aggregates a cost volume computed from input left and right images using 3D convolutions.
    Recent works showed that utilization of extracted image features
    and a spatially varying cost volume aggregation complements 3D convolutions.
    However, existing methods with spatially varying operations are complex, cost considerable computation time, and cause memory consumption to increase. In this work, we construct Guided Cost volume Excitation (GCE) and show that simple channel excitation of cost volume guided by image can improve performance considerably. 
    Moreover, we propose a novel method of using top-$k$ selection prior to soft-argmin disparity regression for computing the final disparity estimate.
    Combining our novel contributions, we present an end-to-end network that we call Correlate-and-Excite (CoEx). Extensive experiments of our model on the SceneFlow, KITTI 2012, and KITTI 2015 datasets demonstrate the effectiveness and efficiency of our model and show that our model outperforms other speed-based algorithms while also being competitive to other state-of-the-art algorithms. Codes will be made available at \url{https://github.com/antabangun/coex}.

\end{abstract}

\section{INTRODUCTION}
\input{1.introduction}

\label{sec.intro}

\section{Related Works}
\input{2.related}

\label{sec.related}

\section{Method}

\input{3.method}

\label{sec.method}

\section{Experiments}

\input{4.experiments}
\label{sec.experiments}

\section{Conclusion}
\input{5.conclusion}

\label{sec.conclusion} 

\addtolength{\textheight}{-0cm}   



\section*{APPENDIX}

\input{appendix}
\label{sec.appendix}



{\small
\bibliographystyle{IEEEtran}
\bibliography{egbib}
}

\end{document}

%% file: def2.tex
\newcommand{\be}{\begin{eqnarray}}
\newcommand{\ee}{\end{eqnarray}}
\newcommand{\bee}{\begin{eqnarray*}}
\newcommand{\eee}{\end{eqnarray*}}

\newcommand{\matrixb}{\left[ \begin{array}}
\newcommand{\matrixe}{\end{array} \right]}

%% file: 1.introduction.tex
Stereo matching aims to estimate depth from a pair of images~\cite{hirschmuller2007stereo,scharstein2002taxonomy} and is an essential task in the field of robotics, autonomous driving, and computer vision.
This task has various challenging issues such as occlusions, textureless areas, areas with repeating textures, thin or small objects, etc. 
With the advancements of deep learning algorithms, the accuracy of stereo matching algorithms has improved significantly; however, many accurate state-of-the-art models do not have fast processing speed for real-time applications~\cite{cheng2018depth,yin2019hierarchical,zhang2019ga,lee2019learning,gu2020cascade}. 
Algorithms that focus on fast computations exist but often sacrifice accuracy to gain this advantage which may be the main reason why stereo cameras are not utilized more frequently in applications~\cite{lee2021learning,zhou2017unsupervised} such as autonomous driving where fast computation is essential. If the efficiency of stereo matching algorithms can be improved from the current standard, stereo camera based depth perception can be an alternative to the expensive LiDAR sensors that are currently used in many self-driving algorithms~\cite{hwang2016fast}.

\begin{figure}[t]
\begin{center}
    \includegraphics[trim={0.cm 1cm 0.1cm 0.7cm},width=1\linewidth]{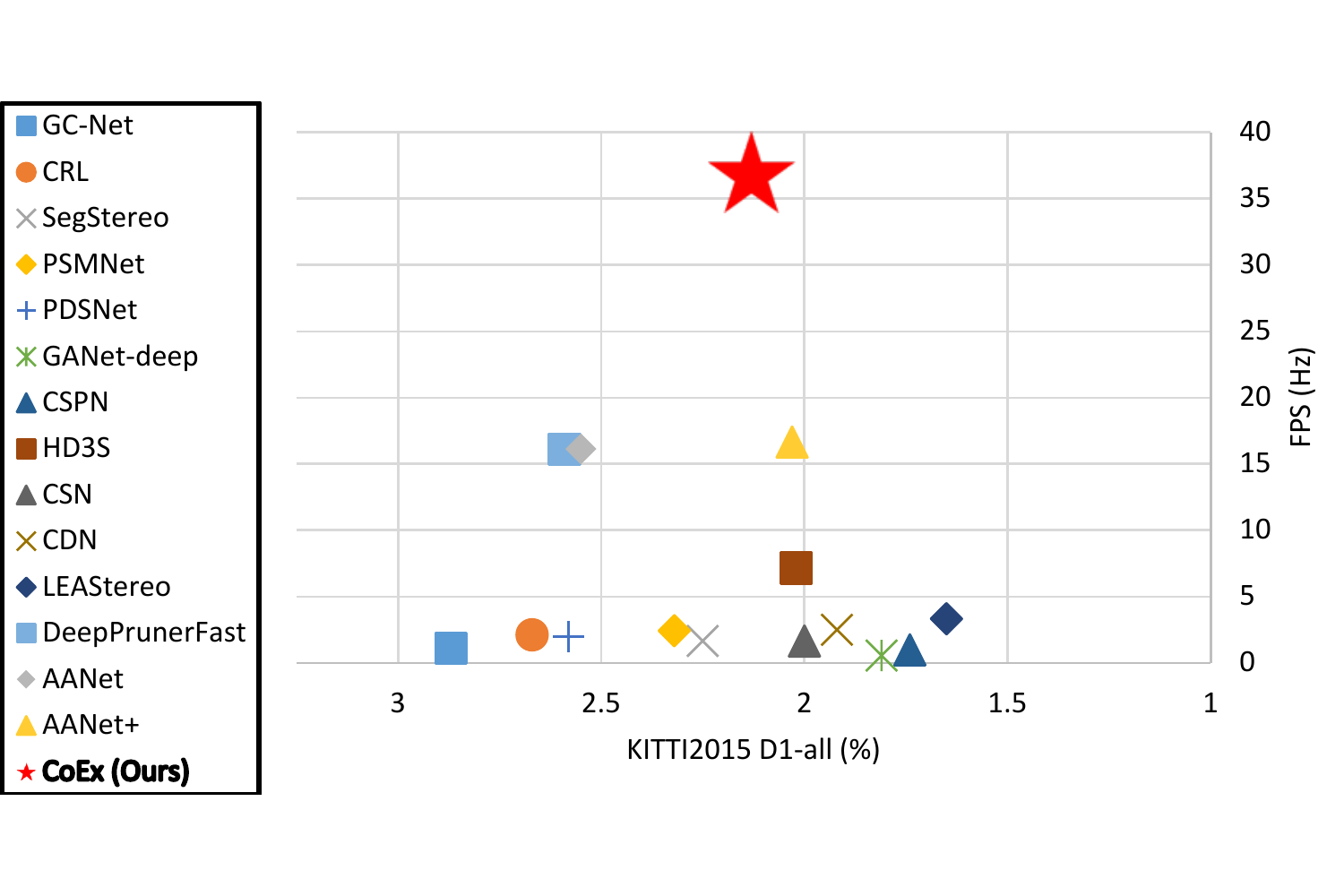}
\end{center}
    \caption{D1-all\% error on KITTI stereo 2015 leaderboard vs. frame rate. Our proposed method CoEx, shown in the red star, achieve competitive performance compared to other state-of-the-art models while also being real-time.}
    \vspace{-7mm}
\label{fig:teaser}
\end{figure}

\begin{figure*}[ht]
\begin{center}
    \includegraphics[trim={0cm 5cm 0cm 0cm}, width=1\linewidth]{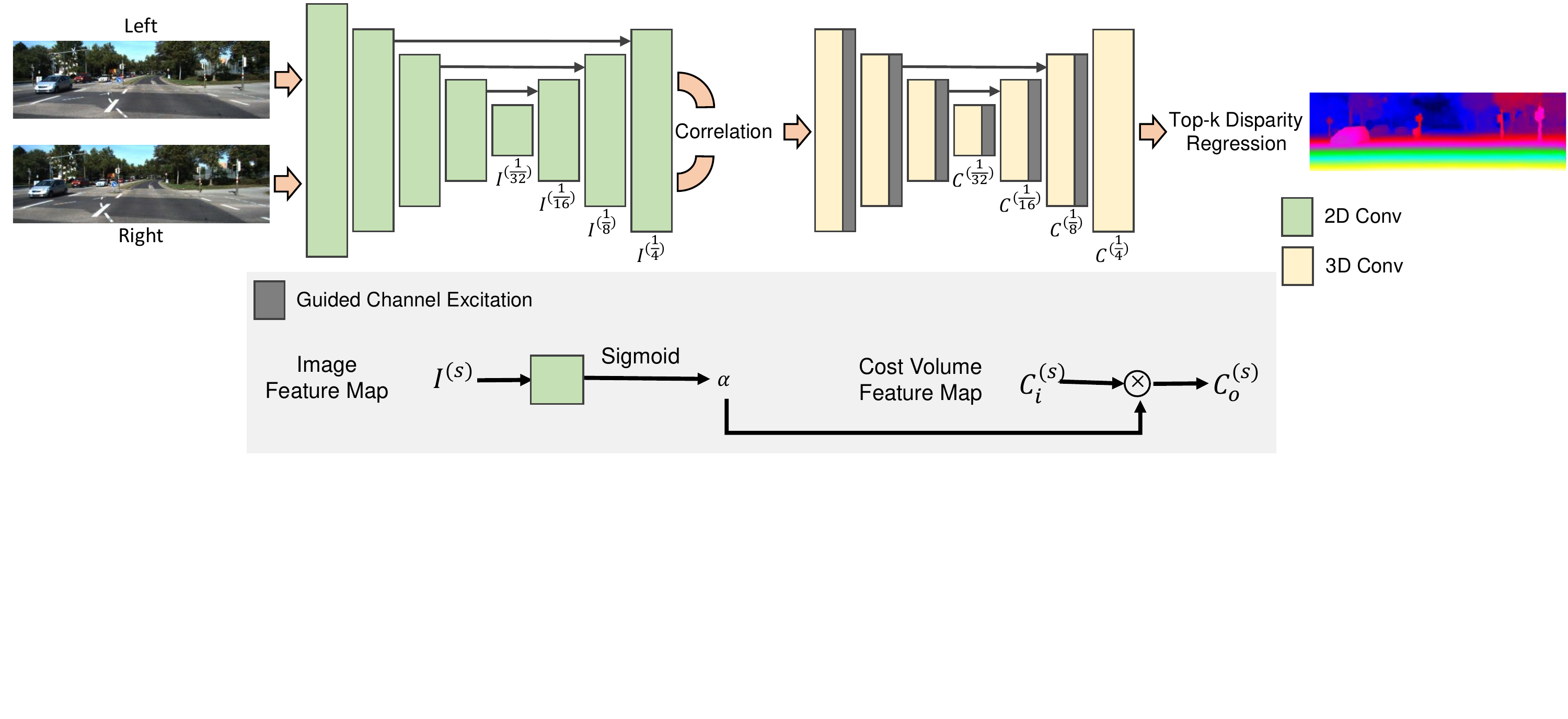}
\end{center}
    \caption{An overall end-to-end stereo matching model with an hourglass architecture of cost aggregation. GCE modules are inserted in between the 3D convolutions to utilize image feature map. Operation between image and cost volume features are broadcasted operation. Top-$k$ regression is used to compute the final disparity estimate. This model can be extended to other volumetric CNN-based architecture and the proposed modules can be incorporated in the same manner.}
    \vspace{-5mm}
\label{fig:overall}
\end{figure*}

Recent series of learning-based stereo matching algorithms~\cite{chang2018pyramid,mayer2016large,zhang2019ga} use left and right input images to construct a cost volume by computing the cross-correlation or concatenation of the features between from the two images. The correlation based approach reduces the input images' feature vectors into cosine similarity values, giving a model with lower memory usage and faster runtime. However, this reduces the representation power of the neural network and often results in poor performance compared to the concatenation based cost volume.

In a volumetric approach, the computed cost volume is aggregated using 3D convolutional layers~\cite{kendall2017end}. However, deep stacks of 3D convolutions are computationally expensive and memory inefficient~\cite{tulyakov2018practical}. 
Recent works have tried to improve the efficiency of the cost aggregation step using spatially varying aggregation ~\cite{zhang2019ga, cheng2018depth, cai2020end}. 
While these works show improvements in accuracy, 
there is a significant increase in computational cost and memory consumption as well as additional complexity in the implementation of the proposed approaches.

We propose an efficient and straightforward way of improving cost aggregation by utilizing extracted image features using attention based approaches that have been shown to improve image classification networks~\cite{woo2018cbam,hu2018squeeze}.
Given a cost volume feature map, Guided Cost volume Excitation (GCE) excites the cost volume channels with weights computed from the the reference image features. The computed weights are shared across the disparity channel, so the operation is lightweight and easy to implement. This module lets the 3D neural network layers to extract geometric features from the cost volume and the image-guided weights to excite the relevant features.
We empirically show that this operation improves performance significantly without any significant additional computational cost.
We show that this module allows correlation based cost volume to utilize image information and performs at a similar accuracy with the concatenation based model, allowing us to construct a fast and accurate correlation based stereo matching model.


In volumetric based stereo matching models, soft-argmin is the standard approach to compute the final disparity estimates, and few works have been done to improve the soft-argmin regression. The soft-argmin function computes the expected value from a disparity distribution at each pixel obtained from the cost volume aggregation. However, in many cases, the disparity distribution can have multiple peaks \eg, on the edge boundaries or even an almost uniform distribution \eg, textureless region. 
Due to this reason, taking the expected value when the distribution is not unimodal may not be the best choice to estimate the disparity. Instead,
we propose to use only the top-$k$ values from the distribution to compute the disparity map. We show that this simple yet novel idea gives more accurate depth estimates and can be applied to any volumetric based model.

With our proposed ideas, we construct an end-to-end real-time stereo matching network that we call CoEx (Correlate-and-Excite).
We sum up our contributions and list them as follows:
\begin{enumerate}
    \item We present Guided Cost volume Excitation (GCE) 
    to utilize extracted feature map from image as guidance for cost aggregation to improve performance. 
    \item We propose a new method of disparity regression in place of soft-argmax(argmin) to compute disparity from the top-$k$ matching cost values and show that it reliably improves performance.
    \item Through these methods, we build a real-time stereo matching network CoEx, that outperforms other speed-oriented methods and shows its competitiveness when compared to state-of-the-art models.
\end{enumerate}

%% file: 2.related.tex

Recent works have focused on using deep Convolutional Neural Networks (CNN) to improve stereo matching performance. In \cite{luo2016efficient,zbontar2016stereo,lee2019visuomotor}, CNNs are used to obtain feature representation for left and right images to be used for feature matching, but cost aggregation is still done using traditional means. DispNet~\cite{mayer2016large} extended the idea to train an end-to-end deep model to predict depth from stereo images by introducing a correlation layer to construct the cost volume. Following this, many more end-to-end works have been proposed which can mostly be divided into either direct regression or volumetric approach~\cite{cheng2020hierarchical}. Direct regression based methods use 2D convolutions on the cost volume to directly compute the disparity map~\cite{pang2017cascade,yang2018segstereo,song2020edgestereo}. On the other hand, volumetric based methods use 3D convolutions to aggregate the cost volume by taking into account the geometric constraints~\cite{chang2018pyramid,kendall2017end,tulyakov2018practical,cheng2020hierarchical} and stacking 3D convolutions in an hourglass architecture. 

Recently, more works have focused on improving the efficiency of 3D convolutions in the aggregation step. Two notable works GANet~\cite{zhang2019ga} and CSPN~\cite{cheng2018depth} use spatially dependent filters to aggregate cost. 
These methods have achieved a higher accuracy using spatially dependent 3D aggregation but
at the cost of a higher computation time. Inspired by the strengths and drawbacks of these approaches, we base our model on spatially dependent 3D operation but focus on speed and efficiency.
On the other hand,
StereoNet~\cite{khamis2018stereonet} focused on building a real-time stereo matching model, and like many others, do so by sacrificing its accuracy. Recently, the accuracy of works~\cite{duggal2019deeppruner,xu2020aanet} on real-time stereo matching models are getting closer to the best performing models.

The volumetric based approaches mentioned above outputs a distribution of matching cost values
at each disparity level for every pixel. The final disparity estimates are then computed by taking the expected value of the distribution using a soft-argmin operation. As a result, the network is only indirectly trained to produce a disparity distribution and can fail in ambiguous regions. There have been few works improving the soft-argmin disparity regression. Recent studies AcfNet~\cite{zhang2020adaptive} and
CDN~\cite{garg2020wasserstein} train the network to produce better unimodal distribution by introducing novel loss functions. This work presents a new method that builds upon the soft-argmin operation itself and improves the overall disparity regression.

%% file: 3.method.tex
A deep learning based end-to-end stereo matching network consists of matching cost computation, cost aggregation, and disparity regression. 
We present a novel GCE and top-$k$ soft-argmin disparity regression module that can be integrated into volumetric based baseline stereo approaches, both without adding significant computation overhead to the baseline stereo matching model. 
A real-time end-to-end stereo model is built using the proposed modules, shown in \figref{fig:overall}, that achieves competitive performance to the state-of-the-art. We describe each of the components in detail in the following subsections.

\subsection{Matching cost computation}
\label{sec.method.matching}

Given a left and right input stereo image pair $3\times H\times W$, feature maps are extracted from both of them using a shared feature extraction module. We use MobileNetV2 \cite{sandler2018mobilenetv2} as our backbone feature extractor for its lightweight property and build a U-Net~\cite{ronneberger2015u} style upsampling module with long skip connections at each scale level. From this feature extraction module, features at each scale are extracted for use later as a guiding signal for spatially varying cost aggregations. 
To construct the cost volume, feature maps extracted at the $1/4$ scale of the left and right image are used with correlation layer~\cite{mayer2016large} to output a $D/4\times H/4\times W/4$ cost volume, where $D=192$ is the maximum disparity set for our network.



\subsection{Guided Cost volume Excitation (GCE)}
3D convolutions are used in modern architectures to aggregate the constructed cost volume to allow the neural network to capture geometric representation from the data. Recent works~\cite{cheng2018learning,zhang2019ga} have used spatially varying modules to complement 3D convolutions and lead to better performance. Specifically, weights are computed from the reference image feature map to aggregate the 3D feature representation computed from the cost volume. The modules compute weights at each location for each pixel of interest and its surrounding neighbors to allow for neighborhood aggregation in a spatially dependent manner. 

We argue that the 3D convolutions in a volumetric cost aggregation already capture neighborhood information. A spatially varying update of the cost volume feature map without neighborhood aggregation is sufficient and is significantly more efficient. To formulate it, for a cost volume with $c$ feature channels, we pass an image feature map at the same scale into a guidance sub-network to output $c$ weights for each pixel.
With this formulation, the 3D convolutions capture geometric information from the cost volume, and the guidance weights excite the relevant geometric features. At scale $(s)$ of the cost volume:


\begin{equation} 
\begin{split}
    \alpha = \sigma(F^{2D}(I^{(s)})) \\
    C_o^{(s)} = \alpha\times C_i^{(s)}, 
    \label{eq:channel}
\end{split}
\end{equation}
where $F^{2D}$ is implemented using 2D point-wise convolution, with $\sigma$ being the sigmoid function. The guidance weights are shared across the disparity dimension, and the multiplication in \eqref{eq:channel} is a broadcasted multiplication. This flow is shown on the bottom left of \figref{fig:overall}. Since this module involves excitation of cost volume features using weights computed from the reference image feature map as guidance, we call this module Guided Cost volume Excitation (GCE). This module is extremely simple and straightforward, with only a few operations added to the overall network; however, we show in ~\secref{sec.exp.abl.gce} that adding GCE module can improve the accuracy of our model significantly. 
In our CoEx model, the cost aggregation architecture follows GC-Net~\cite{kendall2017end}, with an hourglass architecture of 3D convolutions but with a reduced number of channels and network depth to reduce computational cost. The proposed GCE module is then added at every scale of the cost volume (~\figref{fig:overall}). The overall cost aggregation module with GCE is detailed in ~\tabref{tab.costagg}. The module outputs a 4D cost volume at $1/4$ of the original image resolution.

\subsection{Top-$k$ disparity regression}
\label{sec.method.topk}
The 4D cost volume produced in the previous steps gives us matching confidence values for each disparity level for every pixel, which can be transformed into a probability distribution by taking a Softmax across the disparity values. In previous works, the soft-argmax operation is used to compute disparity by taking the expected value over this distribution
\cite{kendall2017end}:
\begin{equation}
    \hat{d}=\sum_{d=0}^{D}d\times Softmax(c_d)
\end{equation}
where $d$ is a predetermined set of disparity indices. 

A disparity distribution where there is only a single peak may give an adequate estimate for disparity predictions. However, in some instances, there can be multiple peaks
or even a relatively uniform distribution.
In these cases, the expected value of the matching cost distribution can diverge significantly from the actual ground truth value. 

To alleviate this issue, instead of taking the expected value of the whole distribution, we use only the top-$k$ 
values of the aggregated cost volume at every pixel. We call this regression strategy top-$k$ soft-argmax(argmin) disparity regression. 
Specifically, at every pixel, we use the top-$k$ weights to compute the expected disparity value. 

When $k$ equals the number of disparity of interest $D$, the top-$k$ regression is simply a soft-argmax operation~\cite{kendall2017end}. When $D>k>1$, only the top-$k$ values in each pixel are used to compute the estimated disparity. This is done by masking the top-$k$ values and performing softmax on these values to normalize them so that weights that sum up to 1 can be obtained. These weights are then multiplied with their corresponding disparity indices, while the remaining values are masked out. The sum of the values are the weighted average of the top-$k$ disparity candidates.
This operation can be seen as similar to $k$-max pooling~\cite{kalchbrenner2014convolutional}. 
In the instance where $k$ equals $1$, the top-$k$ regression becomes an argmax,
since the weight of the maximum index becomes a constant at $1$. When this is the case, the operation is not trainable, and is why previous works resorted to using soft-argmax. 
Though simple, we show through our experiments the effectiveness of the top-$k$ soft-argmax regression. 

Using the top-$k$ regression to compute the disparity map at the full resolution requires a large amount of additional computation time, as shown in \secref{sec.exp.abl}. To mitigate this, we design our model to compute the disparity regression at $1/4$ of the input image resolution.
Finally, the output disparity prediction is upsampled to the original input image resolution. 
Following the footsteps of~\cite{yang2020superpixel}, 
the final disparity estimate at each pixel in the upsampled resolution is obtained with a weighted average of a $3\times 3$ ``superpixel'' surrounding it. Another CNN branch predicts the weights for each superpixel.

We train the network in a fully supervised end-to-end manner using $\textit{smooth}_{L1}$ loss function. Our final loss function is as follows:
\begin{equation}
    \mathcal{L}(d_{GT},\hat{d})=\frac{1}{N}\sum_{i=1}^{N}\textit{smooth}_{L_1}(d_{GT,i}-\hat{d}_i),
\end{equation}
given,
\begin{equation}
    \textit{smooth}_{L_1}(x)=
    \left \{
        \begin{tabular}{ll} 
            $0.5x^2$, & \textit{if} $|x|<1$ \\
            $|x|-0.5$, & \textit{otherwise}
        \end{tabular},
    \right. \\
\label{eq_smooth_l1}
\end{equation}
where, $N$ is the number of labeled pixels, $d_{GT}$ and $\hat{d}$ is the ground truth and predicted disparity respectively.

%% file: 4.experiments.tex
\input{Tables/Comparison_small}

In this section, we explain in detail the implementation details and training of our Correlate-and-Excite (CoEx) network, show through extensive experiments and ablations the effectiveness of our approach, and include detailed discussions on our method.

\subsection{Datasets and Evaluation metrics}

To test the effectiveness of our approach CoEx, we conduct experiments and evaluations on the following datasets: SceneFlow~\cite{mayer2016large}, KITTI Stereo 2012~\cite{geiger2012we}, and KITTI Stereo 2015~\cite{menze2015object}. 

SceneFlow is a synthetic dataset consisting of 35,454 training images and 4,370 testing images. 
The disparity range starts from 1 to 468, with all images having a size of $W=960$, $H=540$. We use the `finalpass' version of the dataset. 
Only pixels with disparity values lower than our maximum disparity of 192 are used for training and evaluation. The end-point-error (EPE), which is the average difference between the predicted and ground truth, is used as a reporting metric. 

KITTI 2012 and 2015 datasets are real-world datasets with sparse ground truth obtained from a LiDAR sensor. 
We divide the training data into 90\% training and 10\% validation set. 
KITTI 2012 uses `Out-All', the percentage of erroneous pixels in total for an error threshold of 3 pixels, for its metric. For KITTI 2015, we show the `D1-all' metric reported on the leaderboard, which is the percentage of all labeled pixels' stereo disparity outliers.

\subsection{Implementation details}

We use the MobileNetV2 pre-trained on ImageNet~\cite{imagenet_cvpr09} as listed in~\secref{sec.method.matching} for our feature extractor backbone. The use of ImageNet pre-trained model allows for faster convergence during training.
We implement our model using PyTorch and use the Adam optimizer ($\beta_1 =0.9$, $\beta_2=0.999$) as our optimizer with Stochastic Weight Averaging (SWA) \cite{izmailov2018averaging}. We randomly crop images to size $W=576$, $H=288$ for training. 

On the SceneFlow dataset, we train our model for 10 epochs with a learning rate $1\times10^{-3}$ for the first 7 epochs and $1\times10^{-4}$ for the remaining 3 epochs with a batch size of $8$. For our experiments on the KITTI dataset, we use a model pre-trained on the SceneFlow dataset and finetune the model on the KITTI dataset for 800 epochs with an initial learning rate of $1\times10^{-3}$ decaying by a factor of $0.5$ at epochs $30$, $50$, and $300$. The Nvidia RTX 2080Ti GPU is used for training and testing. 


\begin{figure*}[t]
\begin{center}
    \includegraphics[trim={0cm 5.3cm 0cm 4.6cm}, clip, width=1\linewidth]{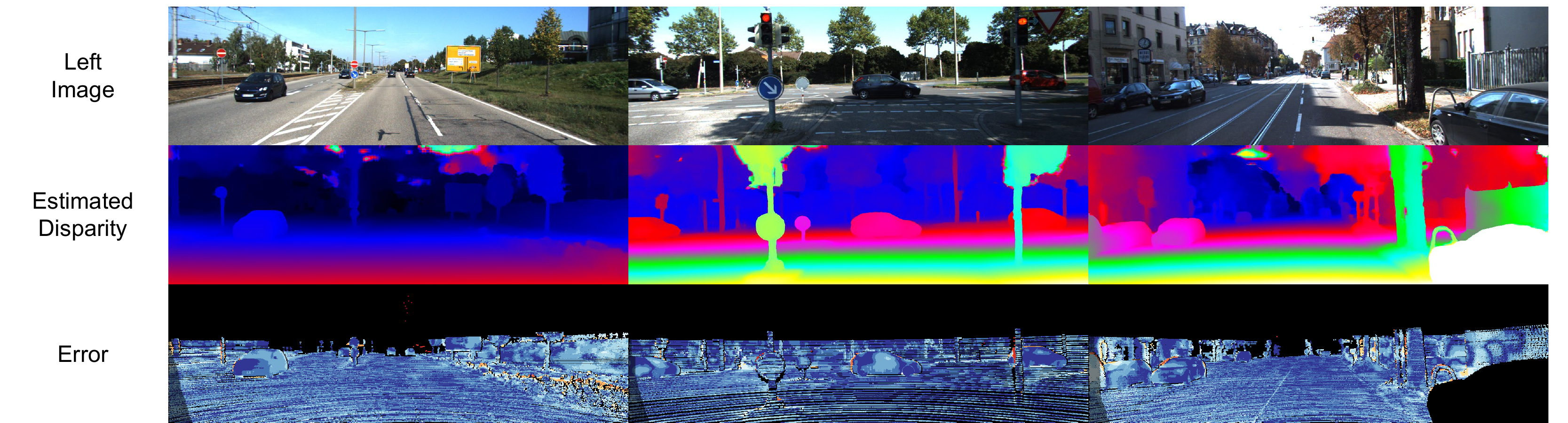}
\end{center}
    \caption{Qualitative results on KITTI 2015 test set. Error in orange corresponds erroneous prediction.}
    \vspace{-4mm}
    \label{fig:kittitest}
\end{figure*}

\begin{figure*}[t]
\begin{center}
    \includegraphics[trim={0cm 5.2cm 0cm 5.2cm}, clip, width=1\linewidth]{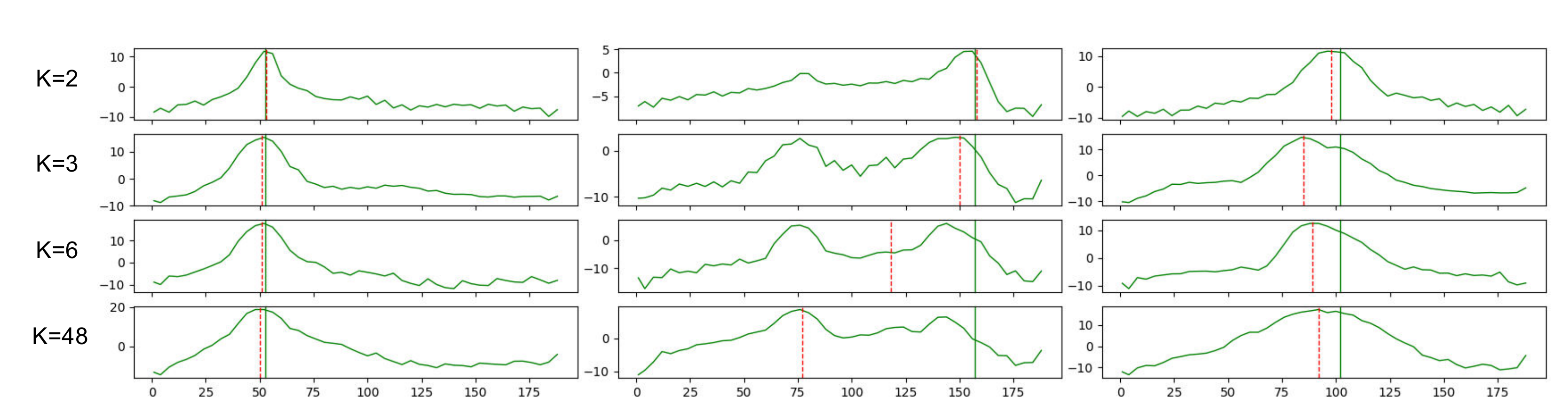}
\end{center}
    \caption{Disparity distributions of models trained with different choice of $k$ in top-$k$ regression. Dashed red line is the estimated disparity and the solid green line is ground truth disparity.}
    \vspace{-5mm}
\label{fig:topk}
\end{figure*}

\subsection{Performance of CoEx}

We show the comparisons of our model to the existing state-of-the-art in \tabref{table:comparison}.
Note that KITTI results are all from the KITTI Stereo Matching Leaderboard, and the SceneFlow EPE values, as well as the runtime, are the values reported in each work. 
Among the speed based models, StereoNet is the fastest performing model with a runtime of 15~$ms$. However, StereoNet's accuracy on SceneFlow and KITTI is considerably less than CoEx,
with differences being 0.411 EPE for SceneFlow and 2.7\% on KITTI 2015. 

\input{Tables/2080Ti_comparison}
As runtime comparisons in different hardware do not give a fair comparison, we compare the runtime breakdown of LEAStereo~\cite{cheng2020hierarchical} and AANet~\cite{xu2020aanet} with our model tested on the same hardware (RTX 2080Ti) using the official open-source models in \tabref{table:2080ticomparison}. The cost aggregation part includes cost volume construction and disparity regression. Our model is $3.3\times$ faster than AANet while giving 0.18 EPE lower and 0.46\% better KITTI 2012 3px out-all\% and 0.42\% better D1-all\% on KITTI 2015. AANet+ added more focus towards disparity refinement to improve accuracy without sacrificing speed at the cost of a high number of network parameters at $8.4M$ compared to our $2.7M$. Our model does not use any post aggregation refinement and still gives similar accuracy while being $3\times$ faster than AANet+.



\subsection{Ablation study}
\label{sec.exp.abl}

\input{Tables/Ablation.all}
\input{Tables/Ablation.add}
\input{Tables/Ablation.graph}
We perform ablation studies on the SceneFlow dataset to study the influence of the proposed modules.
We integrate GCE and top-$k$ soft-argmin regression into baseline stereo matching models. For this ablation study, we used the baseline PSMNet and CoEx model (\tabref{table:all}). Note that PSMNet uses a concatenation of the feature representations between the left and right images to construct cost volume. Concatenation allows the neural network to have a stronger representation power than correlation based cost volume construction that reduces the feature map to a single value of cosine similarity for each match. Replacing the concatenation in PSMNet to correlation reduces the accuracy as expected. However, adding only a single GCE layer into the correlation based PSMNet, indicated by `One' in ~\tabref{table:all}, brings the accuracy to a similar value to the concatenation based PSMNet, indicating that GCE enable the network to utilize image feature representations that is missed by correlation. In addition, the use of correlation also reduces the computation time significantly. 

In PSMNet, the cost volume is upsampled to the original input image resolution and the maximum disparity value is at $D=192$. We test top-$k$ soft-argmin regression in PSMNet with $k$ between $2$ to $192$. We found that reducing $k$ from the original value of $k=192$ generally improves performance up to a point. The accuracy degrades when k is set too low, perhaps due to a lack of gradient flow in backpropagation. Moreover, performing sorting to obtain the top-$k$ values in the full cost volume resolution proves to be too computationally costly.

This motivated us to compute our disparity regression in the CoEx model at $1/4$ the input image resolution and utilize the superpixel upsampling ~\secref{sec.method.topk} to obtain the disparity map at the original resolution.
Note that in CoEx, $k=192/4=48$ is the maximum value of $k$. We show in ~\tabref{table:all}, adding top-$k$ soft-argmin regression to CoEx hardly increases the computation time and gives better accuracy when lower $k$ values are used.

~\tabref{table:all} also shows the performance gain when GCE is integrated at every scale level (\figref{fig:overall}), indicated by `Full'. Our best model is obtained when full GCE integration and top-$2$ soft-argmin regression are added into the base CoEx model. Notice that the two proposed modules only add $1 ms$ of computation overhead from the base model but gives $0.17$ lower test EPE.

\subsubsection{GCE}
\label{sec.exp.abl.gce}

We investigated two approaches to use the reference image as a guide for cost volume aggregation. The first is a simple addition between image features and cost volume features with a broadcasted operation, which effectively acts like a UNet style skip-connection. The second is based on excitation and is the proposed GCE module. The test comparison between the two on the SceneFlow dataset is shown in ~\tabref{table:addvsexcite}. Addition based skip-connection does give a slight accuracy improvement to the baseline. However, we found cost volume excitation a much more effective way of utilizing image features in cost aggregation.

We compare the GCE module that performs spatially varying local aggregation with a similar spatially varying operation that involves neighborhood aggregation. To do this, we formulate a neighborhood as a graph and use graph convolution to aggregate the nodes surrounding the center node of interest, where the graph edges are spatially varying and computed from the reference image feature map. The details of this graph-based aggregation are given in the Appendix. ~\tabref{table:gcevsgraph} shows that a simple excitation of the cost volume feature using a GCE module is performs better and more efficient than the implemented neighborhood spatially independent aggregation.

\subsubsection{Top-$2$ disparity regression}


To further illustrate how top-$k$ regression improves compared to soft-argmin regression, we plot the disparity distribution, produced from the output of cost aggregation, of models trained with each $k$ value. \figref{fig:topk} illustrates 3 cases where a lower $k$ value in top-$k$ regression outperforms the baseline soft-argmin method. In the left most plot, the candidate disparities have a unimodal distribution. The middle case shows when there are 2 possible peaks, and the rightmost case shows the case when the distribution is relatively flat. In all those cases, the model trained using top-$2$ distribution is able to use only the peak matching values and is able to suppress
values far away from the correct matching peak, resulting in a more accurate estimate.

Then how well would models trained with full soft-argmin perform when we replace this regression module with top-$k$ soft-argmin at test time? We provide experimental results for this test in~\tabref{table:all} and found no improvement in the accuracy. The models need to learn to use the top-$k$ soft-argmin regression during training. 


%% file: Tables/Comparison_small.tex
\begin{table}[t]
    \begin{center}
        \begin{adjustbox}{width=0.5\textwidth}
        \setlength{\tabcolsep}{5pt}
        \begin{tabular}{c|c|c|c|c|c}
            \Xhline{4\arrayrulewidth}
            &\multirow{3}{*}{Methods} & Scene- & \multicolumn{2}{c|}{KITTI} & \\
            \cline{4-5}
            & & Flow & 2012 & 2015 & Runtime \\
            & & EPE & 3px(\%) & D1(\%) & ($ms$)\\
            \hline
            \multirow{11}{*}{ \rotatebox{90}{\textit{Accuracy}}}
            &GC-Net~\cite{kendall2017end}           & 2.51              & 2.30              & 2.87              & 900 \\ 
            &CRL~\cite{pang2017cascade}             & 1.32              & --                & 2.67              & 470 \\ 
            &SegStereo~\cite{yang2018segstereo}     & 1.45              & 2.03              & 2.25              & 600 \\ 
            &PSMNet~\cite{chang2018pyramid}         & 1.09              & 1.89              & 2.32              & 410 \\ 
            &PDS-Net~\cite{tulyakov2018practical}   & 1.12              & 2.53              & 2.58              & 500 \\ 
            &GANet-deep\cite{zhang2019ga}           & 0.84              & \underbar{1.60}   & 1.81              & 1,800 \\ 
            &CSPN~\cite{cheng2018depth}             & 0.78              & --                & \underbar{1.74}   & 1,000 \\ 
            &HD$^3$S~\cite{yin2019hierarchical}                        & 0.78              & 1.80              & 2.02              & \textbf{140} \\ 
            &CSN~\cite{gu2020cascade}                            & \textbf{0.65}     & --                & 2.00              & 600 \\ 
            &CDN~\cite{garg2020wasserstein}                            & \underbar{0.70}   & --                & 1.92              & 400 \\ 
            &LEAStereo~\cite{cheng2020hierarchical}                      & 0.78              & \textbf{1.45}     & \textbf{1.65}     & \underbar{300} \\ 
            \hline
            \multirow{6}{*}{ \rotatebox{90}{\textit{Speed}}}
            &DispNetCorr~\cite{mayer2016large}      & 1.68              & 4.65              & 4.34              & 60 \\ 
            &DeepPrunerFast~\cite{duggal2019deeppruner} & 0.97          & --                & 2.59              & 62 \\
            &StereoNet~\cite{khamis2018stereonet}   & 1.101             & 6.02              & 4.83              & \textbf{15} \\ 
            &AANet~\cite{xu2020aanet}               & 0.87              & 2.42              & 2.55              & 62 \\ 
            &AANet+~\cite{xu2020aanet}              & 0.72              & 2.04     & \textbf{2.03}     & 60 \\ 
            \rowcolor{red!7}
            &\textbf{CoEx (Ours)}           & \textbf{0.69}              & \textbf{1.93}              & \underbar{2.13}              & \underbar{27} \\ 
            \Xhline{4\arrayrulewidth}
        \end{tabular}
        \end{adjustbox}
    \caption{Comparison with other state-of-the-arts models. \textbf{Bold}: Best, \underbar{Underscore}: Second best.}
    \vspace{-7mm}
    \label{table:comparison}
    \end{center}
\end{table}

%% file: Tables/2080Ti_comparison.tex
\begin{table}[t]
    \begin{center}
        \begin{adjustbox}{width=0.45\textwidth}
        \setlength{\tabcolsep}{4pt}
        \begin{tabular}{c|c|c|c|c}
            \Xhline{4\arrayrulewidth}
            \multirow{2}{*}{Methods} & Feature & Cost & Refine- & \multirow{2}{*}{Total} \\
            & Extraction & Aggregation & ment & \\
            \hline
            LEAStereo~\cite{cheng2020hierarchical} & 12 & 463 & -- & 475 \\
            AANet~\cite{xu2020aanet} & 22 & 32 & 32 & 88 \\
            AANet+~\cite{xu2020aanet} & 11 & 21 & 45 & 80 \\
            CoEx (Ours) & \textbf{10} & \textbf{17} & -- & \textbf{27} \\
            \Xhline{4\arrayrulewidth}
        \end{tabular}
        \end{adjustbox}
    \caption{Time comparison in $ms$ with other state-of-the-arts models on the same hardware. \textbf{Bold}: Best time.}
    \vspace{-7mm}
    \label{table:2080ticomparison}
    \end{center}
\end{table}


%% file: Tables/Ablation.all.tex
\begin{table}[t]
    \begin{center}
        \begin{adjustbox}{width=0.48\textwidth}
        \begin{tabular}{c|c|c|c|c|c|c}
            \Xhline{3\arrayrulewidth}
            Base & \multicolumn{2}{c|}{Cost volume} & \multirow{2}{*}{GCE} & Top-k reg & SceneFlow & Time \\
            model & Corr & Concat & & $k$ & EPE & (ms)
            \\
            \Xhline{3\arrayrulewidth}
            PSMNet 
                   & & \checkmark & & 192 & 0.8291 & 292 \\
                   & & \checkmark & & 6 & 0.7437 & 405 \\
                   & & \checkmark & One & 192 & 0.8176 & 297 \\
                   & & \checkmark & One & 6 & \textbf{0.7321} & 405 \\
                   & \checkmark & & & 192 & 1.053 & 223 \\
                   & \checkmark & & & 6 & 0.8798 & 332 \\
                   & \checkmark & & One & 192 & 0.8285 & 225 \\
                   & \checkmark & & One & 192$\rightarrow$6* & 0.8088 & 333 \\
                   & \checkmark & & One & 6 & 0.7653 & 333 \\
                   & \checkmark & & One & 2 & 1.108 & 332 \\
            \Xhline{3\arrayrulewidth}
            CoEx 
                  & \checkmark & & & 48 & 0.8552 & 26 \\
                  & \checkmark & & & 6 & 0.8262 & 26 \\
                  & \checkmark & & & 3 & 0.7928 & 26 \\
                  & \checkmark & & & 2 & 0.7942 & 26 \\
                  & \checkmark & & One & 48 & 0.8242 & 26 \\
                  & \checkmark & & Full & 48 & 0.7426 & 26 \\
                  & \checkmark & & Full & 48$\rightarrow$2* & 0.7782 & 27 \\
                  & \checkmark & & Full & 6 & 0.7185 & 27 \\
                  & \checkmark & & Full & 3 & 0.7115 & 27 \\
                  & \checkmark & & Full & 2 & \textbf{0.6854} & 27 \\
            \Xhline{3\arrayrulewidth}
        \end{tabular}
        \end{adjustbox}
    \caption{Ablation study of GCE and top-$k$ soft-argmin regression integrated into base models on SceneFlow `finalpass' with the EPE metric (lower is better). `One' means only a single GCE layer is added into the model, while `Full' adds GCE layer at every scale level~\figref{fig:overall}. \newline *$k_1\rightarrow k_2$ : the model is trained using $k_1$ and tested using $k_2$ soft-argmin regression}
    \label{table:all}
    \vspace{-9mm}
    \end{center}
\end{table}

%% file: Tables/Ablation.add.tex
\begin{table}[t]
    \begin{center}
        \begin{adjustbox}{width=0.33\textwidth}
        \begin{tabular}{c|c|c|c|c}
            \Xhline{3\arrayrulewidth}
            Base & \multicolumn{2}{c|}{GCE} & \multicolumn{2}{c}{SceneFlow} \\
            model & Add & Excite & EPE & 3px \\
            \Xhline{2\arrayrulewidth}
            CoEx   & & & 0.7426 & 4.308 \\
                   & \checkmark & & 0.7310 & 4.159 \\
                   & & \checkmark & 0.6854 & 4.021 \\
            \Xhline{3\arrayrulewidth}
        \end{tabular}
        \end{adjustbox}
    \caption{Comparison between the use of addition or excitation of cost volume features from reference image features.}
    \vspace{-3mm}
    \label{table:addvsexcite}
    \end{center}
\end{table}

%% file: Tables/Ablation.graph.tex
\begin{table}[t]
    \begin{center}
        \begin{adjustbox}{width=0.46\textwidth}
        \begin{tabular}{c|c|c|c|c|c}
            \Xhline{3\arrayrulewidth}
            Base & \multirow{2}{*}{GCE} & \multirow{2}{*}{Neighborhood} & \multicolumn{2}{c|}{SceneFlow} & Time \\
            model & & & EPE & 3px & (ms)
            \\
            \Xhline{2\arrayrulewidth}
            CoEx   & One & & 0.7684 & 4.409 & 26 \\
                   & & One & 0.7732 & 4.435 & 47 \\
            \Xhline{3\arrayrulewidth}
        \end{tabular}
        \end{adjustbox}
    \caption{Comparison of GCE and spatially varying neighborhood aggregation}
    \vspace{-8mm}
    \label{table:gcevsgraph}
    \end{center}
\end{table}

%% file: 5.conclusion.tex
This paper introduces a new real-time stereo matching model that leverages spatially dependent cost aggregation that we call CoEx. We show that spatially varying aggregation can be performed in a lightweight and straightforward fashion to improve performance. We also show how a direct use of top-$k$ values can improve the soft-argmin disparity regression.
We believe that the incredible speed of our method, where it is fast enough for real-time applications, can be a springboard for future real-time stereo matching research in real-world application settings.

%% file: appendix.tex
\subsection{Detailed architecture}
The detailed cost aggregation module is shown in \tabref{tab.costagg}. $s$ and $p$ are stride and padding sizes for the convolution kernels respectively. $I^{(s)}$ is the feature map of the left image obtained in the feature extraction stage at scale $(s)$.
\label{sec.appendix.det}
\input{Tables/costagg}

\subsection{Neighborhood aggregation}
\label{sec.appendix.neigh}

There are multiple previously proposed methods performing spatially varying aggregation that utilizes the neighborhood information \cite{cai2020end,cheng2018learning,zhang2019ga}. To compare GCE with a module that computes spatially varying aggregation of the neighbors, here we formulate a module that performs image-guided neighborhood aggregation. Given a voxel of interest at pixel location $i$ and its neighbors $j\in N(i)$ in a $1\times n\times n$ window, we compute the feature update of cost volume at $i$ as follows:
\begin{equation} 
\begin{split}
    m_{i}^{(s,t+1)}=\sum_{j\in N(i)}e_{ji}\odot C^{(s,t)}_j , \quad\quad\quad\\
    C_{i}^{(s,t+1)}=\xi(W_1C_i^{(s,t)} + W_2m_i^{(t+1)} + b) , \\
    \label{eq:aggmpnn}
\end{split}
\end{equation}
where $\odot$ represent element-wise product and $\xi$ is an activation function. $e_{ji}$ is the edge weight (or affinity in \cite{cheng2018learning} of $j$ to $i$, and it is computed using $MLP$ on the image features at $i$ and $j$, and also the encoding of the relative position $p_i-p_j$ of the neighbors: 
\begin{equation} 
\begin{split}
    \hat{e}_{ji} = MLP([I_i^{(s)}||I_j^{(s)}||MLP(p_i-p_j)]) \\
    e_{ji}^c = \exp{\hat{e}_{ji}^c}/\sum_{j\in N(i)}\exp{\hat{e}_{ji}^c},
    \label{eq:edge}
\end{split}
\end{equation}
 where we use $\text{softmax}$ (2nd line of the eqation) to normalize the edge weights at each feature channel $c$. In this work, Deep Graph Library (DGL) \cite{wang2019dgl} is used to implement the neighborhood aggregation as a graph.

For image feature map with $c_I$ channels and cost volume of size $c\times d\times h\times w$, GCE requires the following computation cost:
\begin{equation} 
\begin{split}
    (c_I\times c\times h\times w) + (c\times d\times h\times w)
    \label{eq:compcostgce}
\end{split}
\end{equation}
Where the left part of the equation is the cost to obtain spatially varying weights, and the right part is the self-update. In contrast, if we write down the cost of weight computation and update of neighborhood aggregation in a $1\times n\times n$ neighborhood, in the simplest form of weight computation where it computes weight by a point-wise convolution, it would require a computation cost of at least:
\begin{equation} 
\begin{split}
   (c_I\times  n\times n\times c\times h\times w) + (n\times n\times c\times d\times h\times w).
    \label{eq:compcostneigh}
\end{split}
\end{equation}
Even in the simplest form, it would require $n\times n$ more times than GCE. 


%% file: Tables/costagg.tex
\begin{table}[h]
    \begin{center}
        \begin{adjustbox}{width=0.45\textwidth}
        \begin{tabular}{cll}
            \Xhline{3\arrayrulewidth}
            \textbf{No.} & \textbf{Layer Setting} & \textbf{Input} \\ 
            \hline
            \hline
            \Xhline{2\arrayrulewidth}
            $[1]$ & correlation layer & $I^{(4)}$ (Left and Right) \\ 
            \Xhline{1\arrayrulewidth}
            $[2$-$1]$ & conv3d $3\times 3\times 3, 8$ & $[1]$ \\ 
            $[2]$ & GCE & $[2$-$1]$ and $I^{(4)}$ \\ 
            $[3$-$1]$ & \bsplitcell{conv3d $3\times 3\times 3, 16$, $s=2$ \\ conv3d $3\times 3\times 3, 16$} & $[2]$ \\ 
            $[3]$ & GCE & $[3$-$1]$ and $I^{(8)}$ \\ 
            $[4$-$1]$ & \bsplitcell{conv3d $3\times 3\times 3, 32$, $s=2$ \\ conv3d $3\times 3\times 3, 32$} & $[3]$ \\ 
            $[4]$ & GCE & $[4$-$1]$ and $I^{(16)}$ \\ 
            $[5$-$1]$ & \bsplitcell{conv3d $3\times 3\times 3, 48$, $s=2$ \\ conv3d $3\times 3\times 3, 48$} & $[4]$ \\ 
            $[5]$ & GCE & $[5$-$1]$ and $I^{(32)}$ \\ 
            
            $[6$-$1]$ & deconv3d $4\times 4\times 4, 32$, $s=2, p=1$ & $[5]$ \\ 
            $[6$-$2]$ & conv3d $3\times 3\times 3, 32$ & $[6$-$1]$ \\ 
            $[6]$ & GCE & $[6$-$2]$ and $I^{(16)}$ \\ 
            $[7$-$1]$ & deconv3d $4\times 4\times 4, 16$, $s=2, p=1$ & $[6]$ \\ 
            $[7$-$2]$ & conv3d $3\times 3\times 3, 16$ & $[7$-$1]$ \\ 
            $[7]$ & GCE & $[7$-$2]$ and $I^{(8)}$ \\ 
            $[8]$ & deconv3d $4\times 4\times 4, 1$, $s=2, p=1$ & $[7]$ \\ 
            \hline
            \Xhline{2\arrayrulewidth}
        \end{tabular}
        \end{adjustbox}
    \caption{Cost aggregation module.}
    \label{tab.costagg}
    \vspace{-5mm}
    \end{center}
\end{table}

%% file: root.bbl
\begin{thebibliography}{10}
\providecommand{\url}[1]{#1}
\csname url@samestyle\endcsname
\providecommand{\newblock}{\relax}
\providecommand{\bibinfo}[2]{#2}
\providecommand{\BIBentrySTDinterwordspacing}{\spaceskip=0pt\relax}
\providecommand{\BIBentryALTinterwordstretchfactor}{4}
\providecommand{\BIBentryALTinterwordspacing}{\spaceskip=\fontdimen2\font plus
\BIBentryALTinterwordstretchfactor\fontdimen3\font minus
  \fontdimen4\font\relax}
\providecommand{\BIBforeignlanguage}[2]{{%
\expandafter\ifx\csname l@#1\endcsname\relax
\typeout{** WARNING: IEEEtran.bst: No hyphenation pattern has been}%
\typeout{** loaded for the language `#1'. Using the pattern for}%
\typeout{** the default language instead.}%
\else
\language=\csname l@#1\endcsname
\fi
#2}}
\providecommand{\BIBdecl}{\relax}
\BIBdecl

\bibitem{hirschmuller2007stereo}
H.~Hirschmuller, ``Stereo processing by semiglobal matching and mutual
  information,'' \emph{{IEEE} Trans. on Patt. Anal. and Mach. Intel.}, 2007.

\bibitem{scharstein2002taxonomy}
D.~Scharstein and R.~Szeliski, ``A taxonomy and evaluation of dense two-frame
  stereo correspondence algorithms,'' \emph{International journal of computer
  vision}, 2002.

\bibitem{cheng2018depth}
X.~Cheng, P.~Wang, and R.~Yang, ``Depth estimation via affinity learned with
  convolutional spatial propagation network,'' in \emph{ECCV}, 2018.

\bibitem{yin2019hierarchical}
Z.~Yin, T.~Darrell, and F.~Yu, ``Hierarchical discrete distribution
  decomposition for match density estimation,'' in \emph{CVPR}, 2019.

\bibitem{zhang2019ga}
F.~Zhang, V.~Prisacariu, R.~Yang, and P.~H. Torr, ``Ga-net: Guided aggregation
  net for end-to-end stereo matching,'' in \emph{CVPR}, 2019.

\bibitem{lee2019learning}
S.~Lee, S.~Im, S.~Lin, and I.~S. Kweon, ``Learning residual flow as dynamic
  motion from stereo videos,'' in \emph{IROS}, 2019.

\bibitem{gu2020cascade}
X.~Gu, Z.~Fan, S.~Zhu, Z.~Dai, F.~Tan, and P.~Tan, ``Cascade cost volume for
  high-resolution multi-view stereo and stereo matching,'' in \emph{CVPR},
  2020.

\bibitem{lee2021learning}
S.~Lee, S.~Im, S.~Lin, and I.~S. Kweon, ``Learning monocular depth in dynamic
  scenes via instance-aware projection consistency,'' in \emph{AAAI}, 2021.

\bibitem{zhou2017unsupervised}
T.~Zhou, M.~Brown, N.~Snavely, and D.~G. Lowe, ``Unsupervised learning of depth
  and ego-motion from video,'' in \emph{CVPR}, 2017.

\bibitem{hwang2016fast}
S.~Hwang, N.~Kim, Y.~Choi, S.~Lee, and I.~S. Kweon, ``Fast multiple objects
  detection and tracking fusing color camera and 3d lidar for intelligent
  vehicles,'' in \emph{URAI}, 2016.

\bibitem{chang2018pyramid}
J.-R. Chang and Y.-S. Chen, ``Pyramid stereo matching network,'' in
  \emph{CVPR}, 2018.

\bibitem{mayer2016large}
N.~Mayer, E.~Ilg, P.~Hausser, P.~Fischer, D.~Cremers, A.~Dosovitskiy, and
  T.~Brox, ``A large dataset to train convolutional networks for disparity,
  optical flow, and scene flow estimation,'' in \emph{CVPR}, 2016.

\bibitem{kendall2017end}
A.~Kendall, H.~Martirosyan, S.~Dasgupta, P.~Henry, R.~Kennedy, A.~Bachrach, and
  A.~Bry, ``End-to-end learning of geometry and context for deep stereo
  regression,'' in \emph{ICCV}, 2017.

\bibitem{tulyakov2018practical}
S.~Tulyakov, A.~Ivanov, and F.~Fleuret, ``Practical deep stereo (pds): Toward
  applications-friendly deep stereo matching,'' in \emph{NeurIPS}, 2018.

\bibitem{cai2020end}
C.~Cai and P.~Mordohai, ``Do end-to-end stereo algorithms under-utilize
  information?'' in \emph{3DV}, 2020.

\bibitem{woo2018cbam}
S.~Woo, J.~Park, J.-Y. Lee, and I.~S. Kweon, ``Cbam: Convolutional block
  attention module,'' in \emph{ECCV}, 2018.

\bibitem{hu2018squeeze}
J.~Hu, L.~Shen, and G.~Sun, ``Squeeze-and-excitation networks,'' in
  \emph{CVPR}, 2018.

\bibitem{luo2016efficient}
W.~Luo, A.~G. Schwing, and R.~Urtasun, ``Efficient deep learning for stereo
  matching,'' in \emph{CVPR}, 2016.

\bibitem{zbontar2016stereo}
J.~Zbontar and Y.~LeCun, ``Stereo matching by training a convolutional neural
  network to compare image patches,'' \emph{Journal of Machine Learning
  Research}, 2016.

\bibitem{lee2019visuomotor}
S.~Lee, J.~Kim, T.-H. Oh, Y.~Jeong, D.~Yoo, S.~Lin, and I.~S. Kweon,
  ``Visuomotor understanding for representation learning of driving scenes,''
  in \emph{BMVC}, 2019.

\bibitem{cheng2020hierarchical}
X.~Cheng, Y.~Zhong, M.~Harandi, Y.~Dai, X.~Chang, H.~Li, T.~Drummond, and
  Z.~Ge, ``Hierarchical neural architecture search for deep stereo matching,''
  in \emph{NeurIPS}, 2020.

\bibitem{pang2017cascade}
J.~Pang, W.~Sun, J.~S. Ren, C.~Yang, and Q.~Yan, ``Cascade residual learning: A
  two-stage convolutional neural network for stereo matching,'' in
  \emph{ICCVw}, 2017.

\bibitem{yang2018segstereo}
G.~Yang, H.~Zhao, J.~Shi, Z.~Deng, and J.~Jia, ``Segstereo: Exploiting semantic
  information for disparity estimation,'' in \emph{ECCV}, 2018.

\bibitem{song2020edgestereo}
X.~Song, X.~Zhao, L.~Fang, H.~Hu, and Y.~Yu, ``Edgestereo: An effective
  multi-task learning network for stereo matching and edge detection,''
  \emph{International Journal of Computer Vision}, pp. 1--21, 2020.

\bibitem{khamis2018stereonet}
S.~Khamis, S.~Fanello, C.~Rhemann, A.~Kowdle, J.~Valentin, and S.~Izadi,
  ``Stereonet: Guided hierarchical refinement for real-time edge-aware depth
  prediction,'' in \emph{ECCV}, 2018.

\bibitem{duggal2019deeppruner}
S.~Duggal, S.~Wang, W.-C. Ma, R.~Hu, and R.~Urtasun, ``Deeppruner: Learning
  efficient stereo matching via differentiable patchmatch,'' in \emph{ICCV},
  2019.

\bibitem{xu2020aanet}
H.~Xu and J.~Zhang, ``Aanet: Adaptive aggregation network for efficient stereo
  matching,'' in \emph{CVPR}, 2020.

\bibitem{zhang2020adaptive}
Y.~Zhang, Y.~Chen, X.~Bai, S.~Yu, K.~Yu, Z.~Li, and K.~Yang, ``Adaptive
  unimodal cost volume filtering for deep stereo matching.'' in \emph{AAAI},
  2020.

\bibitem{garg2020wasserstein}
D.~Garg, Y.~Wang, B.~Hariharan, M.~Campbell, K.~Q. Weinberger, and W.-L. Chao,
  ``Wasserstein distances for stereo disparity estimation,'' \emph{NeurIPS},
  2020.

\bibitem{sandler2018mobilenetv2}
M.~Sandler, A.~Howard, M.~Zhu, A.~Zhmoginov, and L.-C. Chen, ``Mobilenetv2:
  Inverted residuals and linear bottlenecks,'' in \emph{CVPR}, 2018.

\bibitem{ronneberger2015u}
O.~Ronneberger, P.~Fischer, and T.~Brox, ``U-net: Convolutional networks for
  biomedical image segmentation,'' in \emph{MICCAI}, 2015.

\bibitem{cheng2018learning}
X.~Cheng, P.~Wang, and R.~Yang, ``Learning depth with convolutional spatial
  propagation network,'' \emph{arXiv preprint arXiv:1810.02695}, 2018.

\bibitem{kalchbrenner2014convolutional}
N.~Kalchbrenner, E.~Grefenstette, and P.~Blunsom, ``A convolutional neural
  network for modelling sentences,'' \emph{arXiv preprint arXiv:1404.2188},
  2014.

\bibitem{yang2020superpixel}
F.~Yang, Q.~Sun, H.~Jin, and Z.~Zhou, ``Superpixel segmentation with fully
  convolutional networks,'' in \emph{CVPR}, 2020.

\bibitem{geiger2012we}
A.~Geiger, P.~Lenz, and R.~Urtasun, ``Are we ready for autonomous driving? the
  kitti vision benchmark suite,'' in \emph{CVPR}, 2012.

\bibitem{menze2015object}
M.~Menze and A.~Geiger, ``Object scene flow for autonomous vehicles,'' in
  \emph{CVPR}, 2015.

\bibitem{imagenet_cvpr09}
J.~Deng, W.~Dong, R.~Socher, L.-J. Li, K.~Li, and L.~Fei-Fei, ``{ImageNet: A
  Large-Scale Hierarchical Image Database},'' in \emph{CVPR09}, 2009.

\bibitem{izmailov2018averaging}
P.~Izmailov, D.~Podoprikhin, T.~Garipov, D.~Vetrov, and A.~G. Wilson,
  ``Averaging weights leads to wider optima and better generalization,''
  \emph{arXiv preprint arXiv:1803.05407}, 2018.

\bibitem{wang2019dgl}
M.~Wang, D.~Zheng, Z.~Ye, Q.~Gan, M.~Li, X.~Song, J.~Zhou, C.~Ma, L.~Yu,
  Y.~Gai, T.~Xiao, T.~He, G.~Karypis, J.~Li, and Z.~Zhang, ``Deep graph
  library: A graph-centric, highly-performant package for graph neural
  networks,'' \emph{arXiv preprint arXiv:1909.01315}, 2019.

\end{thebibliography}
